# Uncovering Dominant Social Class in Neighborhoods through Building Footprints: A Case Study of Residential Zones in Massachusetts using Computer Vision

Qianhui Liang and Zhoutong Wang

## Abstract

In urban theory, urban form is related to social and economic status. This paper explores to uncover zip-code level income through urban form by analyzing figure-ground map, a simple, prevailing and precise representation of urban form in the field of urban study. Deep learning in computer vision enables such representation maps to be studied at a large scale. We propose to train a DCNN model to identify and uncover the internal bridge between social class and urban form. Further, using hand-crafted informative visual features related with urban form properties (building size, building density, etc.), we apply a random forest classifier to interpret how morphological properties are related with social class.

___________________________________________________________

Q. Liang
77 Massachusetts Avenue, Cambridge, Massachusetts
Email:  qianhuil@mit.edu

Z. Wang (Corresponding author)
48 Quincy Street, Cambridge, Massachusetts
Email: zwang1@gsd.harvard.edu

## 1. Introduction

Winston Churchill says, "we shape our buildings, and afterwards our buildings shape us." Although Churchill is not making any general statement, the philosophy behind this quote is that, buildings and city form are connected to and influence people living in them. In fact, scholars have widely accepted that cities are mirrors and microcosms of society and culture at large, with every viewpoint contributing something to their understanding [1]. For instance, in urban expansion theory, different patterns of urban expansion could be associated with specific environmental costs, such as land consumption and mobility generation [2]. Those costs are related with housing prices, which lead to different market behaviors of different social classes and results in the concentration of dominant classes and exclusion of other classes. Jacobs [3] also observed different parts of New York to conclude that a compact city form, compared to a less dense one will better improve the overall well-being of city dwellers. Another example is that, indefensible urban form[4], e.g. neighborhoods in Five Oaks, Dayton, Ohio, has evoked the transition in homeowners from middle-income to minorities, renters which has results in a substantially declined in property value.

Meanwhile, US communities are growing increasingly homogenized. A recent study [5] suggests that income segregation between neighborhoods within the nation's greater metropolitan areas, such as Boston, increased by an average of 20 percent from 1990 to 2010. Affluent people choose to live in neighborhoods where almost everyone else is affluent, and poor people are driven to concentrate in the same communities . That is to say, rather than having a mixed and ambiguous state, a certain community tends to be occupied by a clear-cut social class. This trend has grown into a insidious problem in the U.S., leading to social and economic issues.

Given that: first, urban forms have strong connections with social attributes; second, in most cases, a neighborhood is occupied by a dominant social class, we hypothesize that it is possible to arrive at the dominant social class in the area by analyzing its corresponding urban form. In this paper, we train a model using deep convolutional neural network architecture to predict income using figure-ground footprint maps of Massachusetts. The performance of the prediction model suggests that footprint image dataset can shed light on the social class dwelling in a certain zip-code area. Through hand-crafted feature engineering, we analyzed the importance of a series of more interpretable morphological properties in predicting social class from images using a random forest model. This methodology provides an explanation on how urban form is related to social class and why other urban forms are outliers. The results are useful in providing new insight into social segregation and can be utilized as a tool for urban studies.

## 2. Literature Review

In recent years, a common representation of urban form extracts the "matrix of the longest and fewest lines", in other words, the "axial map" embeded from urban space. Spatial syntax [6] has been one of the quantitative methods to analysis the spatial configuration by examining this type of representation. It analyzes urban form by translating the representation of line matrices into a graph. This technique is widely used as a tool to examine the social impact of works by urban planners. However, established upon the theory of topology, the representation of the city discards all metric information and is rather limiting [7].

Meanwhile, a simpler representation is urban texture, reflected in a way of figure-ground maps (Figure 1) which shows the pattern of built spaces, open spaces and so on. It has been widely used as a simple but powerful way to distill structural urban information. Some influential scholars including Colin Rowe [8], discussed the inversion of the solid-void relationship as a characteristic of "model cities". Through professional map-reading training, architects and urban designers have the ability to interpret figure-ground maps for analysis and comparison. In the late 1990s, some quantitative analysis was piloted. Richens [9] proposed the utilization of image processing techniques to analyze the figure-ground map representation and Ratti [10] suggested to add height information (DEM) on figure-ground maps using image processing to measure geometric parameters.

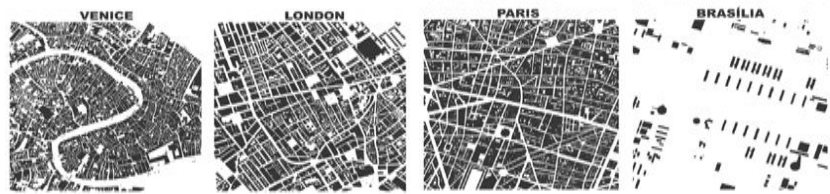

Figure 1. Figure-ground footprint map as a prevailing representation method in urban study

Though informative, creating figure-ground diagrams was a time consuming process a decade ago, and therefore researchers couldn't afford mass production. Hence, it has never been worked with in large scale. Since 2006, the OpenStreetMap has become a welcome source of data as a crowdsourcing platform for figure-ground maps. However, as a Wikipedia for maps, constant quality control is provided by the community itself. Thus, it is hard to verify the quality and completeness of the uploaded maps. Recently paralleled by significant advances in the field of computer vision, Microsoft applied a Deep Convolutional Neural Networks to identify building footprints from Bing satellite imagery and released 124 Million building footprints to the OSM community.

In urban theory, income differences have been considered to be related to spatial characteristic [11]. However, previous social class and segregation research which heavily rely on the Census data [11][12][13], based on a GPS-coded survey and is often limited by time and cost. Differentiating from previous theories that claim figure-ground maps to reflect 'content underneath' of the cities in the field of urban studies, we have three arguments: first, the figure-ground map is a simple but fruitful representation of urban form that can be further utilized to infer social classes. Second, it captures urban form in a more comprehensive manner with a complex set of metrics instead of single aspect. Third, using this method, it is possible to discover which aspects of the visual features of urban form contribute to the interpretation of social class and to what extend.

## 3. Methodology

To answer those questions, we are seeking help from Deep Convolutional Neural Networks (DCNN). DCNN has achieved great success in various computer vision tasks, such as object detection, scene segmentation, etc. DCNN has the ability to capture and interpret visual features, though the reason for its decisions remains unknown.

Arguably, although the accuracy of the DCNN model is very high, it is difficult to understand how the model is making its decisions. To further explain how social classes are inferred from figure-ground maps, we hand-crafted feature engineering algorithms to distill morphological features, e.g., building density, that have specific meanings in urban form. Further, we train a random forest model to fit the image to its corresponding income level. Through analyzing the importance of each variable in the model, we can discover how each feature contributes to the translation.

### *3.1. Data Preprocessing*

In this experiment, for the consistency of income and land use data, we limited the study zone to all residential areas in Massachusetts. We first obtain land-use data from the Massachusetts government website. Subsequently, we select geo-shapes that only correspond to residential zones [14]. We randomly sample 500,000 points in all the residential land-use areas while constraining every point to be at least 80 meters away from one another. Using each sample point as the centroid, we generate figure-ground image patches with the bounding box area of 200m * 200m from the footprint GeoJSON data of Microsoft footprint dataset [15]. Using zip-code level income data from the U.S. census in 2017 [16], we annotate each image with the income levels of the corresponding zip-code which were generated by income figures where its located. The income level is sorted into 8 categories according to U.S. Census survey (Table 1).

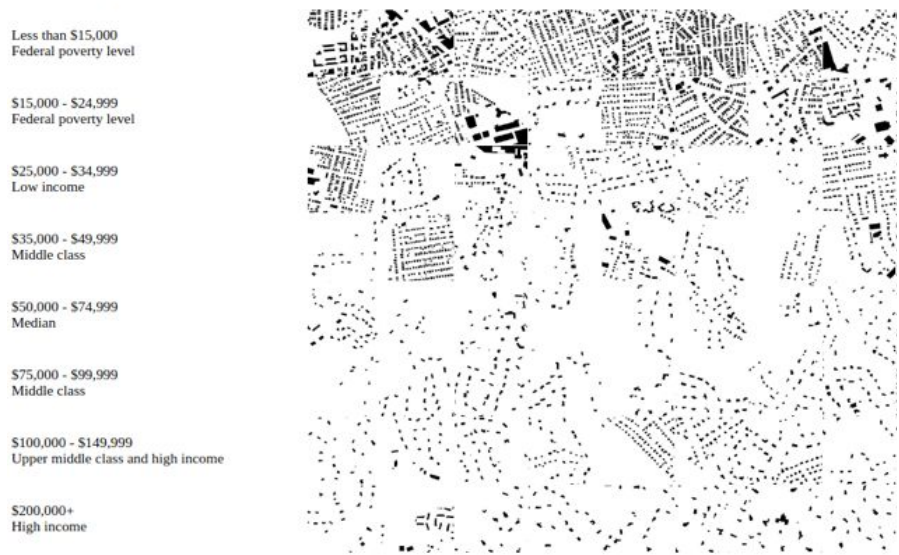

Figure 2. Examples of images from 8 categories

Within each category, the amount of the data is unbalanced. Therefore, to achieve better testing performance, we randomly sample 50000 images from those categories and divided the image corpora into training set, validation set, test set with the ratio of 0.7 : 0.15 : 0.15.

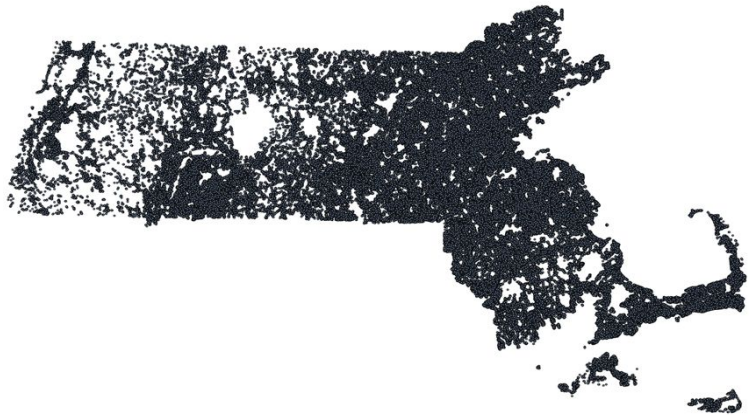

Figure 3. Generated location point within residential area

| Household Income Range | Millions of Household | Percent of Total | Comments | Training Category |
|---|---|---|---|---|
| Less than $15,000 | 14.1 | 11.2% | Federal Poverty Level | 0 |
| $15,000 - $24,999 | 12.1 | 9.6% | | 1 |
| $25,000 - $34,999 | 11.9 | 9.4% | Low Income | 2 |
| $35,000 - $49,999 | 16.3 | 12.9% | Middle Class | 3 |
| $50,000 - $74,999 | 21.5 | 17.0% | Median | 4 |
| $75,000 - $99,999 | 15.5 | 12.3% | Middle Class | 5 |
| $100,000 - $149,999 | 17.8 | 14.1% | Upper Class | 6 |
| $150,000 - $199,999 | 8.3 | 6.6% | High Income | 7 |
| $200,000+ | 8.8 | 7.0% | Obama, Trump High Income | |
| Total | 126.3 | 100% | | |

Table 1. *Household Income Survey, " U.S. Census. and our training category

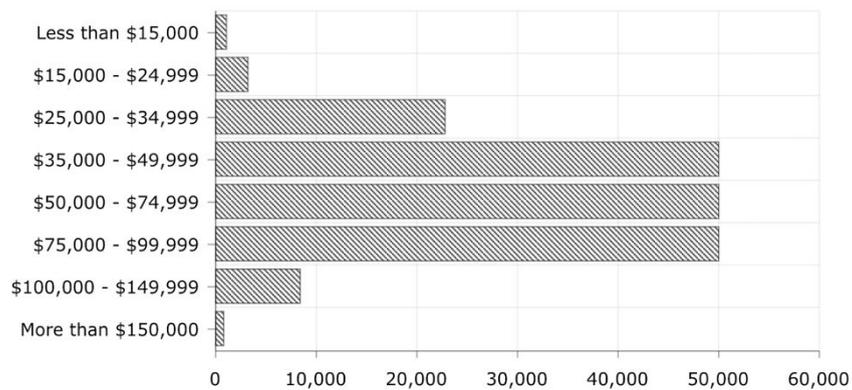

Figure 4. Total number of images collected in each category

### *3.2 Convolutional Neural Networks*

In this paper, we utilize a DenseNet network which connects each layer to every other layer in a feed-forward process. As explained in the paper [17], the main idea that separate it from other DCNN architectures is the $l^{th}$ layer receives the feature-maps of all preceding layers as inputs by concatenating the previous inputs into a single tensor:

$$x_l = H_l([x_0, x_1, \cdots, x_{l-1}])$$

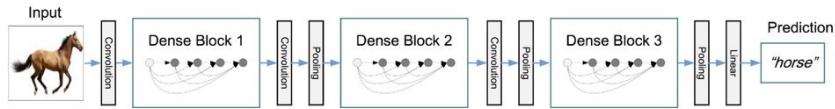

Figure 5. Network structure credit to Huang et al.[17]

First, to interpret DCNN, we adopt the Classification Activation Mapping [18] technique to calculate and visualize discriminative regions. Combining the DenseNet model architecture with the CAM technique can generate a map that shows to what extent different regions contribute to the decision making process.

### 3.3 Feature Engineering and Random Forest

After confirming that the figure-ground maps have strong indications toward income levels using DCNN, , the next step is to find the discriminative feature which can be explained in the relationship between income and pattern configurations.

Then we use hand-crafted feature engineering. We define four categories: building density, building size, contour complexity, directionality. These four categories are chosen according to two criteria: first, they are related with income and housing in the urban theory and second, they are visually extractable from the image. Feature for each category is extracted as a 10 dimension vector.

### 3.3.1 Feature Engineering

**Building density**

Here building density is defined by the coverage of footprint as: $\frac{\sum S_f}{S}$, where $S$ and $S_f$ denotes the total area and building area. In order to capture local density distribution rather than the entire image, we apply sliding windows of different sizes: 224 * 224, 112 * 112, 56 * 56 pixel, to each image (e.g., for a 224*224 image, after applying a 2*2 mask, we get 16 sub-images of size 112 * 112 and 64 sub-images of size 64 * 64). The windows are slided in the image with the stride of half mask length and reflective padding of size stride/2. For each step, we further compute the density of the sub-image of the sliding window. The computation of density for a sub-image is trivial: $density = \frac{number\ of\ black\ pixels}{number\ of\ all\ pixels}$. Eventually, for each category, we calculate the frequency of number of windows that values fall into it. The categories are [0, 10%), [10%, 20%), [20%, 30%), [30%, 40%), {40%, 50%), [50%, 60%), [60%, 70%), [70%, 80%), [80%, 90%), [90%, 100%].

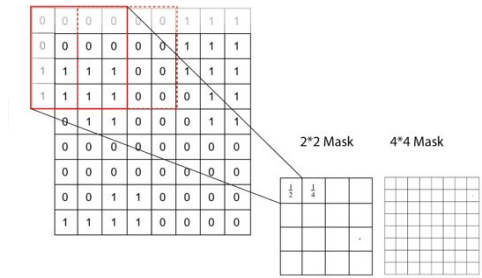

Figure 6. Masks for extract density feature

**Building size**

For each image, we compute the area of every single geometry and sort them into different buckets. We design the range of the buckets roughly according to the distribution of areas of different housing types. The categories are: [0, 50), [50, 100), [100, 150), [150, 200), [200, 250), [250, 300), [300, 350), [350, 40 0), [400, 1000), [1000, +∞). The average footprint area of small single-family houses, duplexes, mid-rise apartments in the U.S are 400ft², 2900ft² and 18200ft², which is 37 m², 269 m² and 1690 m² in international unit [21]. Therefore, most single-family houses, duplexes, mid-rise apartments should fall into the 1st category, 6th and 10th category respectively. For every image, we then compute 10 by 1 vector representing the frequency of numbers of house that falls into 10 categories.

**Contour complexity**

In most of the high-income communities, housings are designed with a more complicated floor plan like adding porches. This will be reflected on the complexity of the footprint contours.

To calculate the contour complexity, we simplified it by measure $\frac{l_c}{\sqrt{S_c}}$. Here, more accurate algorithm can be tested out taking the convex into condition.

**Directionality**

For most of the poor-income communities, the footprint layout is dense and follows the direction of the street. They represent more with such kind of 'grid-like' patterns While in the high-income communities, the directions of houses are more diverse. Here, terrain can be an outlier factors where houses usually follow the direction of the isotypes line. For simplification, we simply ignore this factor.

The directionality is calculated by implementing the image with (Histogram of Oriented Gradients) HOG. Using kernels in Figure 7, the horizontal and vertical gradient is calculated. Next, we can find the magnitude and direction of gradient using the formula. By $g = \sqrt{g_x^2 + g_y^2}$ and $\theta = arctan\frac{g_x}{g_y}$, we can compute the direction of the gradient. Further, we sort out a histogram with the frequency of pixels that fall into category of [0-18, 18-36, 36-54, 54-72, 72-90, 90-108, 108-126, 126-144, 144-162, 162-180]. Notably, only the edges of the buildings are qualified to be taken into account to the histogram.

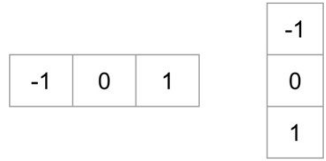

Figure 7. Kernels for HOG

| direction0 (°) | direction1 (°) | direction2 (°) | direction3 (°) | direction4 (°) | direction5 (°) | direction6 (°) | direction7 (°) | direction8 (°) | direction9 (°) |
|---|---|---|---|---|---|---|---|---|---|
| [0, 18) | [18, 36) | [36, 54) | [54, 72) | [72, 90) | [90, 108) | [108, 126) | [126, 144) | [144, 162) | [162, 180) |
| density0 (%) | density1 (%) | density2 (%) | density3 (%) | density4 (%) | density5 (%) | density6 (%) | density7 (%) | density8 (%) | density9 (%) |
| [0.0, 0.1) | [0.1, 0.2) | [0.2, 0.3) | [0.3, 0.4) | [0.4, 0.5) | [0.5, 0.6) | [0.6, 0.7) | [0.7, 0.8) | [0.8, 0.9) | [0.9, 1.0) |
| area0 (m²) | area1 (m²) | area2 (m²) | area3 (m²) | area4 (m²) | area5 (m²) | area6 (m²) | area7 (m²) | area8 (m²) | area9 (m²) |
| [0, 50) | [50, 100) | [100, 150) | [150, 200) | [200, 250) | [250, 300) | [300, 350) | [350, 400) | [400, 1000) | [1000, +∞) |
| complexity0 | complexity1 | complexity2 | complexity3 | complexity4 | complexity5 | complexity6 | complexity7 | complexity8 | complexity9 |
| [3, 6) | [6, 9) | [9, 12) | [12, 15) | [15, 18) | [18, 21) | [21, 24) | [24, 27) | [27, 30) | [30, +∞) |

Table 2. Feature variable explanation

For each image, we engineer feature representations by picking the above four key aspects (direction, building density, building area, contour complexity), and normalizing across these aspects by separating each feature range into 10 discrete intervals, and representing each feature by the probability that each image instance's values fall into each prospective discrete interval as a 10 dimensional vector. The four vectors are then concatenated into a 40 dimensional vector to represent each image instance.

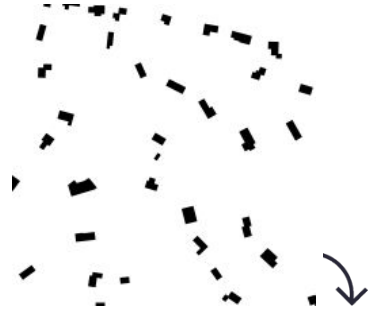

| direction0 | direction1 | direction2 | direction3 | direction4 | direction5 | direction6 | direction7 | direction8 | direction9 |
|---|---|---|---|---|---|---|---|---|---|
| 0.077 | 0.062 | 0.081 | 0.054 | 0.072 | 0.336 | 0.071 | 0.107 | 0.063 | 0.071 |
| density0 | density1 | density2 | density3 | density4 | density5 | density6 | density7 | density8 | density9 |
| 0.067 | 0.121 | 0.105 | 0.231 | 0.331 | 0.008 | 0.137$ | 0.0 | 0.0 | 0.0 |
| area0 | area1 | area2 | area3 | area4 | area5 | area6 | area7 | area8 | area9 |
| 0.294 | 0.647 | 0.029 | 0.029 | 0.0 | 0.0 | 0.0 | 0.0 | 0.0 | 0.0 |
| complexity0 | complexity1 | complexity2 | complexity3 | complexity4 | complexity5 | complexity6 | complexity7 | complexity8 | complexity9 |
| 0.088 | 0.058 | 0.323 | 0.323 | 0.176 | 0.029 | 0.0 | 0.0 | 0.0 | 0.0 |

Figure 8. An example of representation of images by feature engineering: original image (top image) and 40 embedded feature (bottom image).

### *3.3.2 Random Forest Model*

Random Forest is an ensemble learning method [19] for classification and regression tasks. It generates a series of decision trees and counts the majority of their votes to make a final decision. Random Forest uses a modified tree learning algorithm that selects, at each candidate split in the learning process, a random subset of the features. Random forests can be used to rank the importance of variables [20] in a classification problem. The process of fitting the out-of-bag error for each data point is recorded and averaged over the forest. In order to measure the importance of the j-th feature after training, the values of the j-th feature are permuted among the training data and the out-of-bag error is again computed on this perturbed dataset.

## 4. Result

### 4.1. Convolutional Neural Network

We implement the DenseNet 121 on PyTorch framework and train the model using stochastic gradient descent with a batch size of 64 for 100 epochs. The training images are resized into 80 * 80 pixels to fit into GPU memory. The learning rate is initialized as 0.1 and reduced to 1/10 every 30 epochs. The model achieve 87% accuracy overall on the test set.

| Less than $15,000 | $15,000 - $24,999 | $25,000 - $34,999 | $35,000 - $49,999 |
|---|---|---|---|
| 93.27% | 94.18% | 93.28% | 88.00% |
| $50,000 - $74,999 | $75,000 - $99,999 | $100,000 - $149,999 | Higher than 150,000 |
| 80.63% | 92.10% | 88.07% | 91.61% |

Table 3. Prediction accuracy within each categories

### 4.2. Random Forest

We use those manually engineered 40 dimension features of each image to train a random forest classifier to classify the income level for each image. The model achieves 56% accuracy on average. And the overall average weight for each feature is calculated (Figure 9).

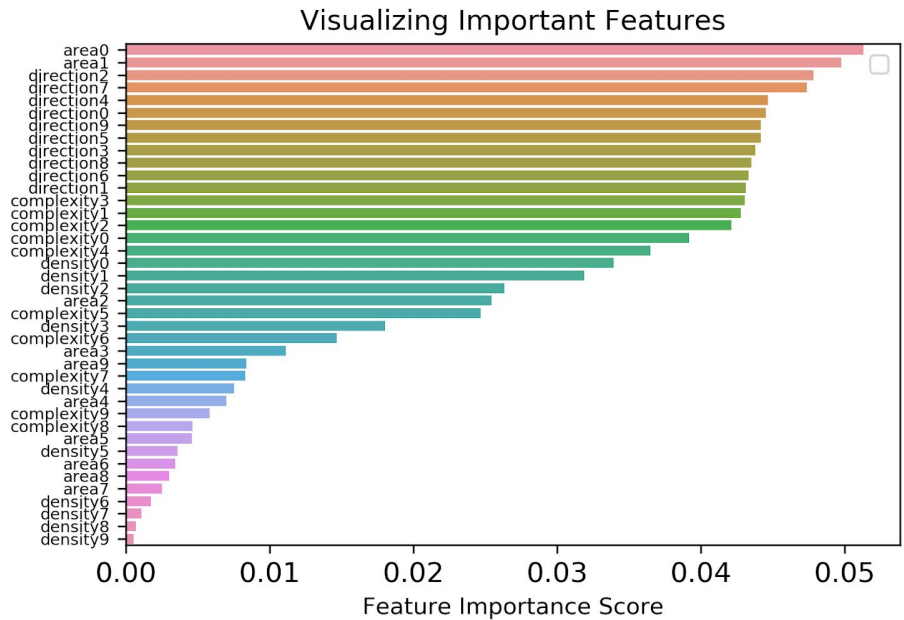

Figure 9. Importance of different features

| Density | Building size | Contour complexity | Directionality |
|---|---|---|---|
| 0.1369 | 0.1632 | 0.2581 | 0.4416 |

Table 4. Important feature

Among all these four categories: building density, building size, contour complexity and directionality, the directionality is more determinant than other features. Against intuition, though density usually considered to be a significant feature to distinguish different social class in common sense, it is less important compared to other geometric properties from the calculation. The appearance and amount of small footprint area buildings (encoded as "area0" and "area1") are informative in inferring the dominant social class.

## 5. Other application of the research

This approach can be used to predict the income level of the missing data area at zip code level resolution and provide new insight into the social segregation.

Meanwhile, we want to identify whether some typical urban form (pattern) exists within each class. Taking use of the feature embedding vector from the last fully-connected layer in the CNN model, we cluster the images within each class (Figure 10) and we can find some obvious clustered pattern representing some typical residential form (Figure 11).

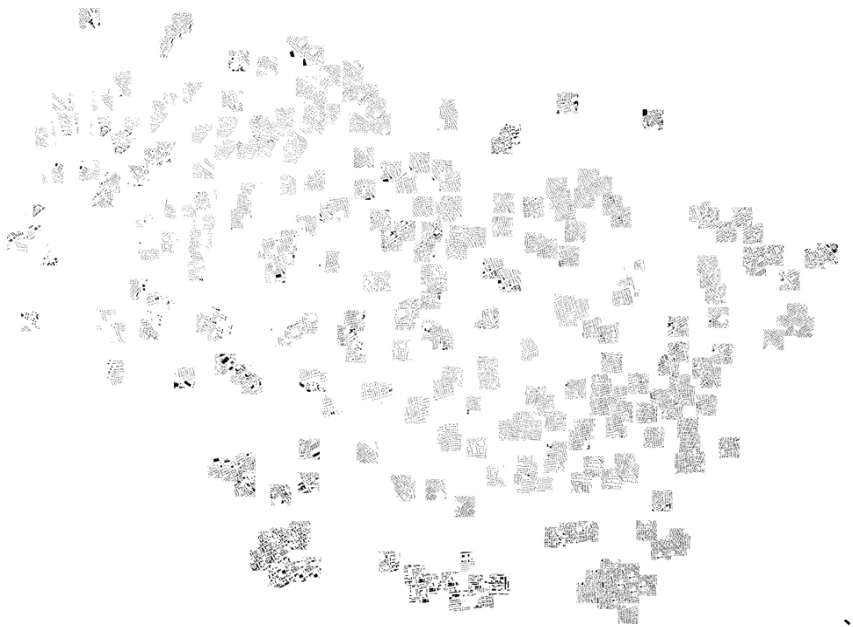

Figure 10. Clustering of images using CNN embedding(class 0)

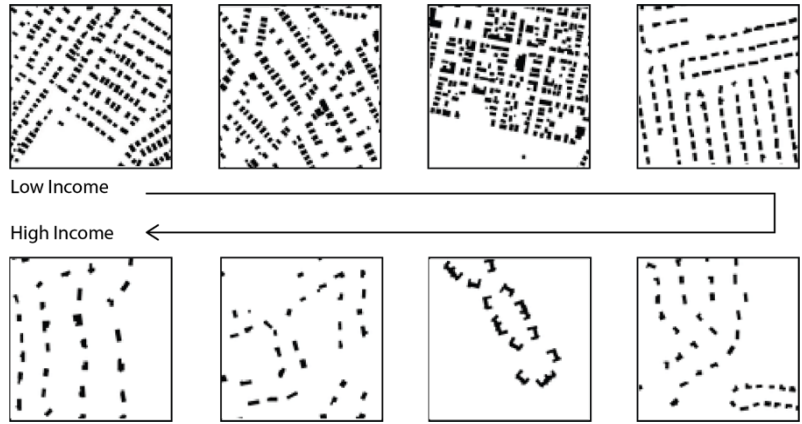

Figure 11. Typical footprint typology identified by the model

## 6. Discussion

In this paper, we introduced a methodology to uncover social class through training a DCNN model on figure-ground footprint maps and a random forest model on manually engineered features from the footprint maps. Our preliminary result provides an alternative approach, in addition to the conventional analytic framework, in predicting social class: rather than the traditional data sources such as demography or housing properties, we utilize simple but informative figure-ground maps.

The major contribution of the work is twofold. First, we have validated the connection between urban forms and social classes through DCNN models Our model achieved 87% accuracy in testing set, indicating that the characteristics of figure-ground maps can capture and predict the dominant social classes. Second, we have proposed a methodology empowered by computer vision to analyze and interpret figure-ground maps, which was only been investigated by professionals in the realm of urban design and planning. The method is scalable and can be generalized into a broader application.

Although the proposed method provides clear value and novel perspectives to the existing research, it also has several limitations. Meanwhile, many potential possibilities can be developed on top of this paper. First, it is worth trying other representations, e.g., figure-ground maps with DEM (a grayscale image containing height information), instead of the binary (black-and-white) figure-ground maps. Moreover, in terms of the feature engineering, spectrums generated by fourier transformation of such binary images may also be informative. Besides, our sample size is limited. In future studies, we are scaling the methodology to more states and validate its performance. Lastly, the scope of our study is only confined to interpreting social class from footprints without analyzing the mechanism of its occurrence. It is necessary to combine local insights to explain prediction results.